\documentclass{article}
\usepackage{ijcai21}
\usepackage{multirow}
\usepackage{subfigure}
\usepackage{times}

\usepackage{soul}
\usepackage{url}
\usepackage[hidelinks]{hyperref}
\usepackage[utf8]{inputenc}
\usepackage[small]{caption}
\usepackage{graphicx}
\usepackage{amsmath}
\usepackage{color}
\usepackage{booktabs}
\usepackage{algorithm}
\usepackage{algorithmic}
\title{CIMON: Towards High-quality Hash Codes}

\author{
Xiao Luo$^{1,2,}$\thanks{Equal Contribution. This work was done when Xiao Luo and Daqing Wu interned in Alibaba Group.  $^\dagger $Corresponding authors.}
\and
Daqing Wu$^{1,2,*}$\and
Zeyu Ma$^{3}$\and
Chong Chen$^{1,2,\dagger}$\and
Minghua Deng$^{1,\dagger}$\and \\
Jinwen Ma$^1$\and 
Zhongming Jin$^2$\and
Jianqiang Huang$^2$\And
Xian-Sheng Hua$^2$
\affiliations
$^1$School of Mathematical Sciences, Peking University, China\\
$^2$DAMO Academy, Alibaba Group, Hangzhou, China\\
$^3$School of Computer Science and Technology, Harbin Institute of Technology, Shenzhen, China
\emails
\{xiaoluo,wudq\}@pku.edu.cn,
zeyu.ma@stu.hit.edu.cn, \{dengmh,jwma\}@math.pku.edu.cn,
\{cheung.cc,zhongming.jinzm, jianqiang.hjq,xiansheng.hxs\}@alibaba-inc.com 
}
\begin{document}

\maketitle

\begin{abstract}

Recently, hashing is widely used in approximate nearest neighbor search for its storage and computational efficiency. Most of the unsupervised hashing methods learn to map images into semantic similarity-preserving hash codes by constructing local semantic similarity structure from the pre-trained model as the guiding information, i.e., treating each point pair similar if their distance is small in feature space. However, due to the inefficient representation ability of the pre-trained model, many false positives and negatives in local semantic similarity will be introduced and lead to error propagation during the hash code learning. Moreover, few of the methods consider the robustness of models, which will cause instability of hash codes to disturbance. In this paper, we propose a new method named  {\textbf{C}}omprehensive s{\textbf{I}}milarity {\textbf{M}}ining and c{\textbf{O}}nsistency lear{\textbf{N}}ing (CIMON). First, we use global refinement and similarity statistical distribution to obtain reliable and smooth guidance. Second, both semantic and contrastive consistency learning are introduced to derive both disturb-invariant and discriminative hash codes. Extensive experiments on several benchmark datasets show that the proposed method outperforms a wide range of state-of-the-art methods in both retrieval performance and robustness.
\end{abstract}

\section{Introduction}
% Nearest neighbor search (NNS) has attracted ever-increasing attention in numerous fields of application, such as computer vision, recommendation systems, cluster analysis and many other machine learning problems\cite{wang2017survey,luo2020survey}. 
% NNS aims to find the closest points from the given dataset for a given query based on a certain distance criteria. 
% NNS methods can be divided into exact methods and approximation methods. When the dataset is large and has high dimensions, it can be typically time-consuming to pinpoint the exact closest data points for a given query. 
% Among various searching algorithms, recently some researchers have paid more attention to approximate nearest neighbor (ANN) search methods, by which the computing complexity can be greatly reduced and the search needs can be properly met as well in most cases. 
Hashing-based Approximate Nearest Neighbour search has attracted ever-increasing attention in the era of big data due to their high retrieval efficiency and low storage cost. The main idea of hashing methods is to project high dimensional datapoints into compact binary codes while preserving the semantic similarity of original datapoints.
% As an effective ANN search solution, hashing-based methods have attracted increasing attention. 
% Based on the powerful representation capabilities of deep learning\cite{lecun2015deep}, more and more deep hashing methods are being proposed, which help to achieve end-to-end hash codes learning outcomes. 

Hashing methods include supervised hashing \cite{luo2020survey,fan20deep,wang2020set,xie2020label} and unsupervised hashing. 
However, it is difficult to apply supervised hashing methods in practice since large-scale data annotations are unaffordable. 
To address this problem, several deep unsupervised methods were proposed and provided a cost-effective solution to practical applications \cite{lin2016learning,yang2018semantic,ijcai2020-479}.
% including deep auto-encoder hashing\cite{krizhevsky2011using}, deep binary descriptors (DeepBit)\cite{lin2016learning}, semantic structure-based unsupervised deep hashing(SSDH)\cite{yang2018semantic}, unsupervised deep hashing by distilling data pairs(DistillHash)\cite{yang2019distillhash} and MLS$^3$RDUH\cite{ijcai2020-479}.
Recently, most unsupervised hashing methods employ a two-step framework: Firstly, the local semantic similarity structure is reconstructed from the pre-trained neural network. To be specific, the local semantic similarity relationships are often derived from the Euclidean distance or the cosine similarity of deep features extracted from the pre-trained model. Secondly, a hashing network is optimized to generate compact and similarity-preserving hash codes by incorporating the defined similarity structure as the guiding information.
\begin{figure}
\centering
\includegraphics[width=8cm,keepaspectratio=true]{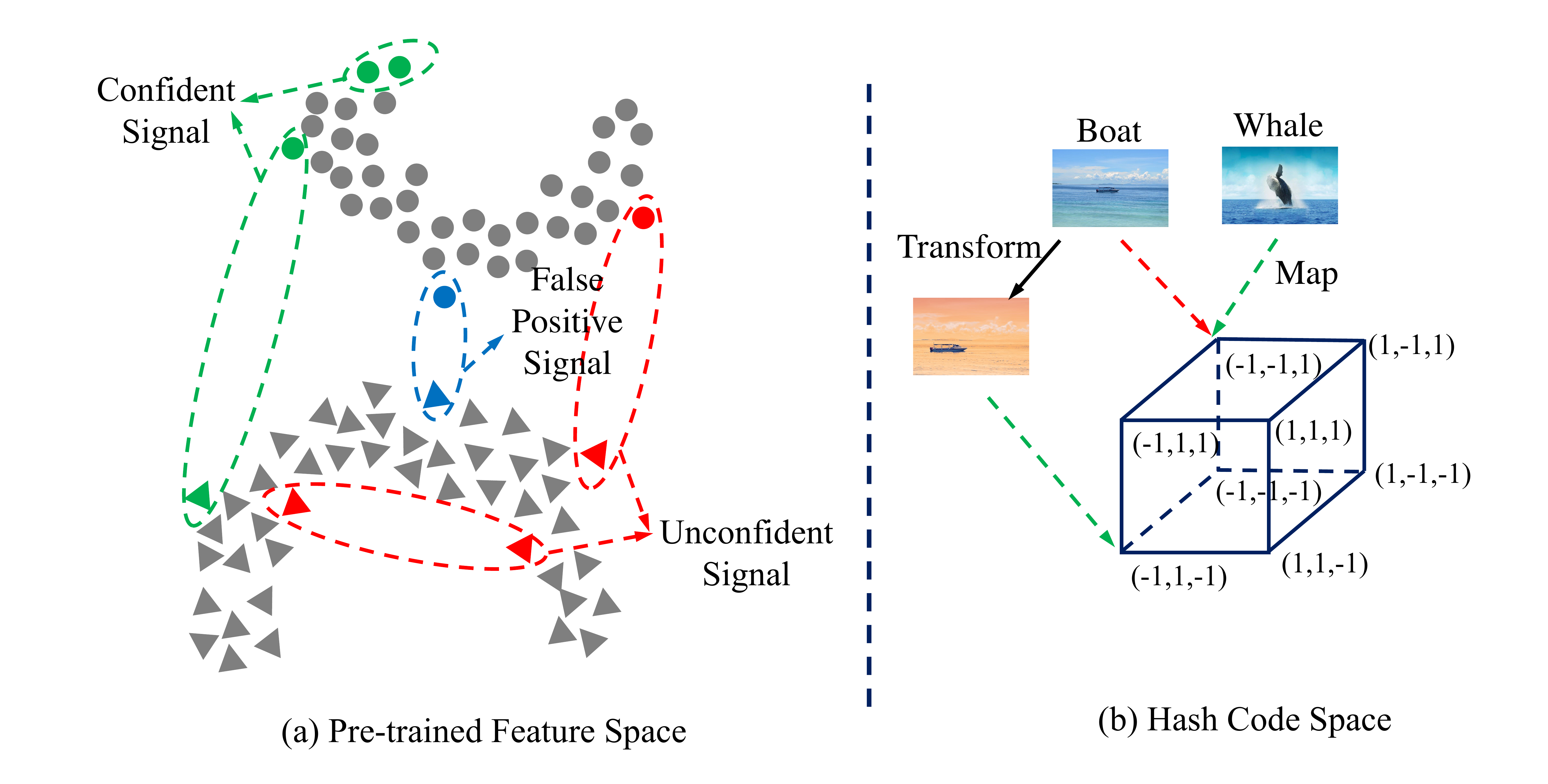}
\caption{Motivation of our model. (a) The “triangle” points and the “circle” points are belong to different categories. False signals (blue) and unconfidence signals (red) will misguide the hash code learning. (b) Two different images (the first line) are mapped to the same hash code (collision) and the hash code is sensitive to the transformation, which implies the hash code is of low quality.}
\label{fig:motivation}
\end{figure}

However, the existing methods have two significant drawbacks that will harm the quality of hash codes. First, many false positives and negatives will be introduced in the similarity matrix for the insufficient representation ability of the pre-trained model, which will misguide the hashing model during hash code learning and further reduce the retrieval performance. As shown in Figure \ref{fig:motivation}(a), false similar pairs can occur between the boundary points of two manifolds (blue points). Moreover, most methods treat the confident signals and uncertain signals equally (green and red points), which will also accumulate a lot of errors. Second, few of them consider the stability of models, which will cause hash codes to be unstable under disturbances and greatly influence the quality. For example, images of different classes with similar background could be mapped to the same hash code (called hash collisions) while the transformed image could be quite far away from the original image in the hash code space (Figure \ref{fig:motivation}(b)).

To address the above two issues, we propose a new method named CIMON, which comprehensively explores semantic similarity structure to achieve reliable semantic guidance and considers the stability of hash codes by introducing consistency learning. Specifically, CIMON firstly takes advantage of the global information to remove false positives between boundary points and smooths the unconfident signals by confidence adjustment. Secondly, CIMON generates two groups of deep features by data augmentation and constructs two similarity matrices and both parallel semantic consistency and cross semantic consistency are encouraged to generate robust hash codes. Furthermore, contrastive consistency between hash codes is adopted to improve the quality of hash codes. Through these improvements, CIMON could obtain high-quality hash codes in both retrieval performance and robustness, which is also demonstrated by extensive experiments on several challenging benchmark datasets. Our main contributions can be summarized as following:

\begin{itemize}
	\item CIMON not only utilizes global refinement to refine the initial local semantic similarity structure, but also explores the similarity statistical distribution to adjust the weight for each image pair, which generates a reliable and smooth guide for hash code learning.
	
% 	CIMON utilizes similarity statistical distribution and global constraint learning to reconstruct the semantic similarity structure to generate a novel similarity pseudo-graph with confidence-based weight. 
	\item A novel consistency loss including semantic consistency and contrastive consistency is proposed to optimize the hashing network, which helps to generate robust and discriminative hash codes.
% 	\item We discuss the relationship between hashing characteristics and contrastive learning. To our knowledge, we are the first to introduce contrastive learning to hash learning.
	
	\item Experiments on several public datasets demonstrate that CIMON outperforms existing state-of-the-art unsupervised hashing techniques by a large margin.
\end{itemize}

\section{Related Work}
% {\bf{Deep Supervised Hashing.}} Deep supervised hashing methods aims to learn deep neural networks that map the data points into Hamming space where the semantic similarities can be preserved\cite{luo2020survey}. CNNH\cite{xia2014supervised} is the first supervised deep hashing method, which is based on a convolutional neural network and performs the hash learning in two steps. The first step is to learn hash codes, and the second step is to learn a specific deep network to map the input samples to the learned hash codes. Typically, Deep Supervised Hashing\cite{liu2016deep} trains a loss function with product form based on the pairwise similarities and Hamming distances by using end-to-end back propagation algorithm. The purpose of the specific loss function is to ensure that pairs of data points with high similarities in the original space have relatively small distance in the Hamming space and pairs with low similarities in the original space have relatively large distance in the Hamming space. Deep Supervised Discrete Hashing\cite{li2017deep} and Deep Pairwise Supervised Hashing\cite{li2016feature} both take advantage of pairwise loss with a Bayesian framework. Deep Triplet Supervised Hashing\cite{wang2016deep} and Deep Neural Network Hashing\cite{lai2015simultaneous} adopt the triplet ranking loss to preserve the relative image similarities. Deep Cauchy hashing\cite{cao2018deep} utilizes Cauchy distribution, which makes the hash codes of samples in the same class all concentrated in a small Hamming ball.

 \paragraph{Deep Unsupervised Hashing.}
Most of the unsupervised deep hashing methods extract deep features to construct a semantic structure, by which unsupervised problems can be turned into supervised problems. In a quite different way, DeepBit \cite{lin2016learning} regards the original images and the corresponding rotated images as similar pairs and attempts to preserve the similarities when learning related hash codes. Stochastic generative hashing \cite{dai2017stochastic} tries to learn hash functions by using a generative model based on the minimum description length principle. SSDH \cite{yang2018semantic} makes use of a specific truncated function on the pairwise distances and constructs the similarity structures. The hashing model is then trained by supervised hashing techniques. Afterwards, the performance of SSDH is improved by DistillHash, which distills the image pairs with confident similarity signals. Clustering-driven Unsupervised Deep Hashing \cite{2019Clustering} recursively learns discriminative clusters by soft clustering model and produces binary code with high similarity responds. MLS$^3$RDUH \cite{ijcai2020-479} reconstructs the local semantic similarity structure by taking advantage of the manifold structure in feature space, achieving the state-of-the-art performance.

 \paragraph{Contrastive Learning.} \cite{hadsell2006dimensionality} is the first work to learn representations by contrasting positive pairs against negative pairs. To solve the storage of large scale dataset, \cite{wu2018unsupervised} proposes to utilize a memory bank for class representation vectors. Various pretext work is based on several forms of contrastive loss function, which is related to the exemplar-based task and noise-contrastive estimation \cite{dosovitskiy2014discriminative}. Recently, Momentum Contrast \cite{he2020momentum} proposes to build a dynamic dictionary with a queue and a moving-averaged encoder, which enables building a large and consistent dictionary on-the-fly that facilitates contrastive unsupervised learning. SimCLR \cite{chen2020simple} further simplifies the learning algorithms without requiring specialized architectures or a memory bank and achieves better performance on ImageNet.

\section{The Proposed Model}
\begin{figure*}[ht]
\centering
\includegraphics[width=17cm,keepaspectratio=true]{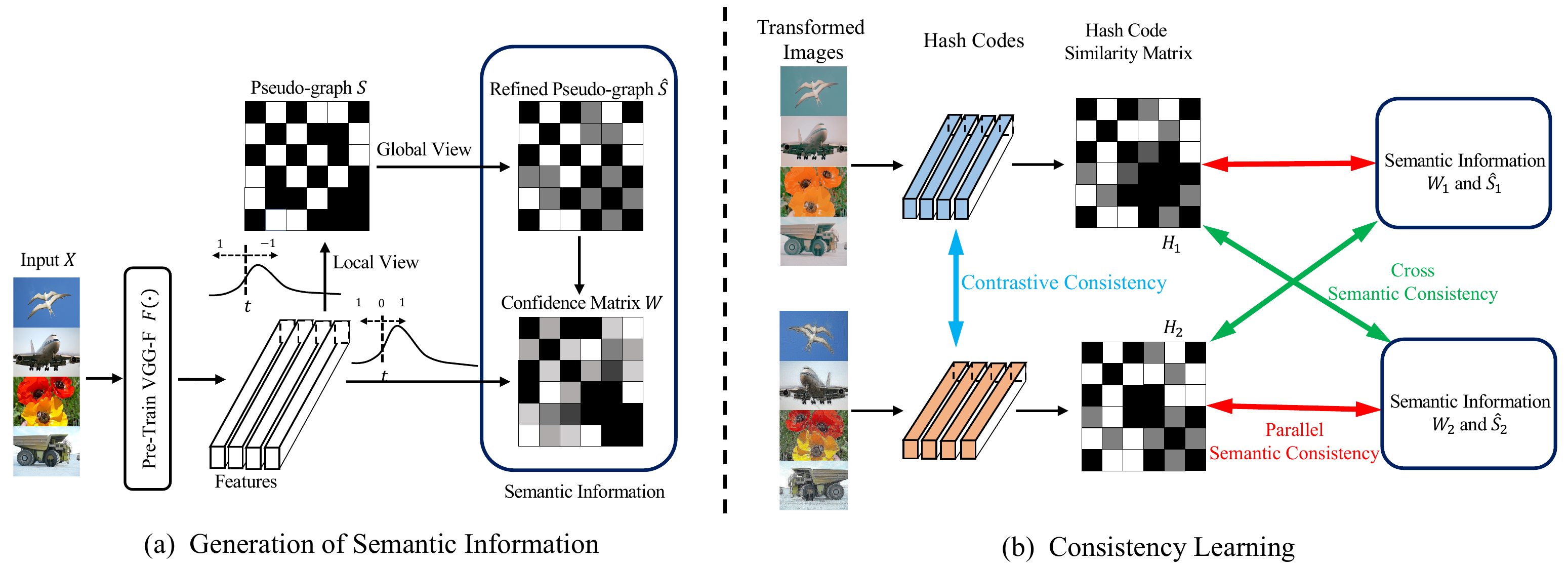}
\caption{Overview of CIMON. (a) CIMON generates semantic information including refined pseudo-graph and confidence matrix for reliable guidance. (b) With data augmentation, CIMON generates two groups of semantic information.  The parallel and cross semantic consistency is constructed between features and hash codes under the same group and crossing the different groups respectively. The contrastive consistency is based on two groups of hash codes. 
% (a) Pseudo-graph is constructed initially from local semantic structure and then is refined by global contraint learning. Moreover, confidence matrix is constructed to reduce the threats of false positives and negatives. (b) With data augmentation, we generate two groups of semantic information. The parallel and cross semantic consistency is constructed between features and hash codes under the same group and crossing the different groups respectively. The contrastive consistency is based on two groups of hash codes to improve the robustness and alleviate the collisions of hash codes. 
}

\label{fig:framwork}
\end{figure*}
In this section, we first formally define the problem and feature our model with two parts as shown in Figure \ref{fig:framwork}: (1)
\textbf{Semantic information generating}. A pre-trained VGG-F \cite{simonyan2014very} without the last fully-connected layer $F(\cdot)$ is adopted to extract deep features, which will be used to generate the similarity graph and the confidence-based weight matrix.
(2) \textbf{Consistency learning}. The hashing network $G(\cdot)$ is modified from VGG-F by replacing the last fully-connected layer with a fully-connected layer with $L$ hidden units to incorporate the hash code learning process. A novel consistency learning framework is adopted to learn high-quality hash codes.
% With data augmentation, we generate two groups of semantic information.
% The parallel and cross semantic consistency is constructed between features and hash codes under the same group and crossing the different groups respectively. The contrastive consistency is based on two groups of hash codes to improve the robustness and alleviate the collisions of hash codes.

\subsection{Problem Formulation}
In the unsupervised hashing settings, $\mathcal{X}=\{x_i\}_{i=1}^N$ is the training set with $N$ samples without label information. We aim to learn a hash function $\mathcal{H}: x\rightarrow b \in\{-1,1\}^L,$
in which $x$ is the input image and $b$ is a compact $L$-bit hash code. This map should preserve similarity, i.e., images with similar ground truth labels should correspond to hash codes with small Hamming distances.

\subsection{Semantic Information Generating}
In our model, semantic information is composed of the similarity pseudo-graph and the similarity confidence matrix.

From the local perspective, the pseudo-graph aims to capture pairwise similarity information. Given the pre-trained deep features $\{ F (x_i) \}_{i=1}^N$, the cosine distance between the $i$-th and the $j$-th images is obtained by $d_{ij}= 1 - \frac{F (x_i)\cdot F (x_j)}{||F (x_i)||_2||F (x_j)||_2}$. Since most pairs should be negative in ground-truth, we set a relatively small threshold $t=0.1$ following \cite{wu2019deep}, and consider pairs with the cosine distance smaller than $t$ as potential positives ($S_{ij}=1$) and pairs with the cosine distance larger than $t$ as potential negatives ($S_{ij}=-1$). Mathematically, the pseudo-graph $S$ can be constructed as:
\begin{equation}
    S_{ij}=\left\{\begin{array}{ll}
    1 & d_{ij}\leq t,  \\ 
    -1 & d_{ij}> t\end{array}\right.
\label{equ:similarity_1}
\end{equation}

 \paragraph{Global Refinement} As mentioned in Figure \ref{fig:motivation}, the features of images with the same semantic information should lay on a high-dimensional manifold following \cite{yang2019distillhash,ijcai2020-479} and many false positives and negatives will be introduced in local semantic similarity $S$ \cite{luo2021}. Hence, we propose to use the global clustering results to refine the semantic similarity by removing contradictory results. Since spectral clustering has been proven to be suitable for clustering high-dimensional manifold data, we take advantage of it to perform global refinement.
%Then we introduce a global constraint to refine the semantic similarity via approximate optimization. Naturally, semantic similar structure should satisfy three essential rules: reflexivity, symmetry and transitivity. The first two rules obviously hold. However, transitivity doesn't always hold in pseudo-graph $S$ (e.g. $S_{ij}=S_{jk}=1$, $\nRightarrow$ $x_i$ is similar to $x_k$). Therefore, we refine pseudo-graph $S$ as $\hat{S}$, which is estimated by finding the most similar matrix to $S$ under transitivity rule. In formulation, 
%\begin{equation}
%\begin{aligned}
%\label{equ:refined}
%\min_{\hat{S}} \quad & \sum_{i, j}\left|\hat{S}_{i j}-{S}_{i j}\right| \\
%\mbox{s.t.}\quad &\hat{S}_{ii}=1, \hat{S}_{ij}=\hat{S}_{ji} &\quad& \forall i, j\\
%                          &g(\hat{S}) = True 
%\end{aligned}
%\end{equation}
%in which $g(\hat{S}) = True$ implies $\hat{S}$ meets the transitivity rule: $S_{ij}=S_{jk}=1$, $\Rightarrow S_{ik}\geq 0$, $\forall i, j, k$.
%
%It is unrealistic to achieve the optimal solution of Equation \ref{equ:refined} since it is NP-hard. However, we can obtain an approximate solution by taking advantage of global information from spectral clustering \cite{NIPS2004_2619}. 
Specifically, assume $c_{i}\in \{1, ..., K\}, i=1,\cdots, N$ is the $i$-th cluster label of spectral clustering ($K$ is the number of clusters). Then each pair of two points with the same class is considered as global potential similar and vice versa. If an image pair has a different potential similarity signal with pseudo-graph $S$, its similarity is considered as unreliable. 
After removing the unreliable signals, the final refined pseudo-graph is $\hat{S}$ is formulated as:
\begin{equation}
   \hat{S}_{ij}=\left\{\begin{array}{ll}
   1 & c_i = c_j \ \& \ S_{ij}=1  \\ 
   -1 & c_i \neq c_j \ \& \ S_{ij}=-1 \\
   0 & \text{otherwise}\end{array}\right. 
\label{equ:similarity_2}
\end{equation} 
%Note that spectral clustering starts from the pairwise similarity and preserves most of the local semantic information while satisfying the global constraint. Specifically, $\hat{S}$ meets the conditions in Equation \ref{equ:refined} and is consistent with the original $S$. Moreover, it is easy to check that the false positives and negatives between the boundary points can be removed after our global refinement.

 \paragraph{Confidence Adjustment} Note that the semantic confidence of similarity signal for each pair is different, we further construct the confidence matrix for pseudo-graph $S$ based on the semantic confidence.
Inspired by recent works \cite{yang2018semantic}, we observe that
the distribution of cosine distances for deep feature pairs can be estimated by two half Gaussian distributions, denoted as $\mathbf{N}(m_1,\sigma_1^2) $ and  $\mathbf{N}(m_2,\sigma_2^2) $ respectively, in which $m_1$, $m_2$ and $\sigma_1^2$ and $\sigma_2^2$ are the corresponding means and variances. 
% More specifically, we consider the image pair $(x_i, x_j)$, of which the distance is smaller than $d_l$ , as a semantically similar one. Following the suggestion in \cite{yang2018semantic}, the distance threshold $d_l$ can be simply denoted as $d_l = (m_1 - 2\sigma _1) $. Meanwhile, we consider image pair $(x_i,x_j)$ with the distance larger than threshold $d_r$ as semantically dissimilar, where dd is denoted as $d_r = (m_2 + 2\sigma_2)$. 
Intuitively, we hypothesize that image pairs with distances obviously smaller than others are semantically ‘similar’ and those with obviously larger distances are semantically ‘dissimilar’, which are denoted as highly confident pairs. Moreover, the confidence weights are approximated by Cumulative Distribution Function (CDF) of two half Gaussian distributions. Specifically, for potential similar signals, distances have a more confident similarity signal if $d_{ij}$ is closer to $0$, and for potential dissimilar signals, distances have a more confident similarity signal if $d_{ij}$ is closer to $2$. 
%According to the theory of confidence interval, we set the weights for confident pairs to 1, whose corresponding distances are far away from the peak. To be specific, we set two distance thresholds $d_l$ and $d_r$, and distances between $d_l$ and $d_r$ have a lower confidence. Note that for the fuzzy pairs ($\hat{S}_{ij}=0$), the weight is set to 0.
% According to the theory of confidence interval, we set two distance thresholds $d_l$ and $d_r$ satisfying $d_l<t<d_r$ and estimate the probability that local semantic similarity accord with the ground truth. 
% Distances have a confident similarity signal out of interval $[d_l,d_r$], distances have a lower confidence between $d_l$ and $d_r$. 
Therefore, confidence-based weight matrix is computed as following:
\begin{equation}
    W_{ij}=\left\{\begin{array}{ll}
    \frac{\Phi_1(t)-\Phi_1(d_{ij})}{\Phi_1(t)-\Phi_1(0)} & d_{ij}\leq t \ \& \ \hat{S}_{ij}\neq 0, \\ 
    \frac{\Phi_2(d_{ij})-\Phi_2(t)}{\Phi_2(2)-\Phi_2(t)} & t < d_{ij} \ \& \ \hat{S}_{ij}\neq 0, \\
    % 1 & d_{ij}\leq d_l\mid d_{ij}\geq d_r \ \& \ \hat{S}_{ij}\neq 0, \\
    0 & \hat{S}_{ij}=0\end{array}\right.
\label{equ:confidence}
\end{equation}
in which $\Phi_k(x)=\Phi(\frac{x-m_k}{\sigma_k})$ and $ \Phi(\cdot) $ is the CDF of standard normal distribution. 
Under this setting, all $W_{ij}\in [0, 1]$ and the confidence weights on both ends are relatively larger.

\subsection{Consistency Learning}
In order to preserve the similarity structure of input images, similar (dissimilar) images are expected to be mapped into similar (dissimilar) hash codes. Different from previous models, here we adopt two groups of semantic information under two different kinds of data augmentation.

 \paragraph{Semantic Consistency} For each image $x_i$, there are two transformed samples $x_i^{(1)}$ and $x_i^{(2)}$. At the semantic information generating stage, two refined similarity graphs with confidence matrices $\{{W}^{(1)},\hat{S}^{(1)}\}$, $\{{W}^{(2)},\hat{S}^{(2)}\}$ are generated with extracted features $\{F(x_i^{(1)})\}_{i=1}^N$ and $\{F(x_i^{(2)})\}_{i=1}^N$ as the guiding information. Simultaneously, images $x_i^{(1)}$ and $x_i^{(2)}$ are the inputs of the hashing network $G(\cdot) $, and hash codes $b_i^{(1)}$ and $b_i^{(2)}$ are obtained through activation function $sign(\cdot)$. Therefore, we derive two similarity outputs $H^{(1)}$ and $H^{(2)}$ from hash codes, which is formulated as 
\begin{equation}
H_{i j}^{(m)}=\frac{1}{L} b_{i}^{(m)\top} b_{j}^{(m)}, \  \ b_{i}^{(m)} = sign(G(x_i^{(m)}; \Theta))
\label{equ:hash_code_similarity}
\end{equation} 
in which $ m = 1 \ \text{or} \ 2$, and $\Theta$ represents the set of parameters of hashing network. For the purpose of preserving the semantic structures, we first minimize weighted $L_2$ loss between the hash code similarity and the corresponding pseudo-graph from the same group. The parallel semantic consistency loss can be formulates as:
\begin{equation}
    \begin{aligned}
        \mathcal{L}_{PSC} & =\frac{1}{N^2}\sum_{i=1}^N\sum_{j=1}^N W_{ij}^{(1)}(H_{ij}^{(1)} - \hat{S}_{ij}^{(1)})^2 \\
        & + W_{ij}^{(2)}(H_{ij}^{(2)} - \hat{S}_{ij}^{(2)})^2
    \end{aligned}
    \label{equ:parallel}
\end{equation}
Inspired by the cross-attention mechanism \cite{boussaha2019multi}, we also match the hash code similarity with the pseudo-graph from the different groups. To be specific, the cross semantic consistency loss can be written as:
\begin{equation}
    \begin{aligned}
        \mathcal{L}_{CSC} & =\frac{1}{N^2}\sum_{i=1}^N\sum_{j=1}^N W_{ij}^{(1)}(H_{ij}^{(2)}
        - \hat{S}_{ij}^{(1)})^2 \\
        &+ W_{ij}^{(2)}(H_{ij}^{(1)} - \hat{S}_{ij}^{(2)})^2
    \end{aligned}
    \label{equ:cross}
\end{equation}

 \paragraph{Contrastive Consistency} Self-supervised learning has been proved to generate high-quality representations for downstream tasks \cite{he2020momentum}.
From this points, we randomly sample a minibatch of $M$ images, producing $2M$ random transformed images $\{x_i^{(1)}, x_i^{(2)}\}_{i=1}^M$. Given a positive pair $x_i^{(1)}$ and $x_i^{(2)}$, we treat the other $2(M - 1)$ augmented images within a minibatch as negative examples. Denote $b_{i}\star b_{j}$ as cosine similarity of $b_i$ and $b_j$ ($\frac{b_i \cdot b_j}{||b_i||_2||b_j||_2}$), the contrastive consistency loss is defined as
\begin{equation}
    \ell_{CC} = -\frac{1}{2M}\sum_{i=1}^{M}\Big(\log \frac{e^{b_i^{(1)}\star b_i^{(2)} / \tau}}{Z_i^{(1)}} + \log \frac{e^{b_i^{(1)}\star b_i^{(2)} / \tau}}{Z_i^{(2)}}\Big)
    \label{equ:contrastive}
\end{equation}
where
$
    Z_i^{(m)} = \sum_{j\neq i}(e^{b_i^{(m)}\star b_j^{(1)} / \tau} + e^{b_i^{(m)}\star b_j^{(2)} / \tau}),
$
$m = 1 \ \text{or} \ 2$, and $\tau$ denotes a temperature parameter set to $0.5$ following \cite{chen2020simple}. Note that the numerator of each term punishes the distance between hash codes of samples under different transformation while the denominator encourages to enlarge the distance between hash codes of different samples, which encourages the hash codes to be uniformly distributed in the hash code space from \cite{pmlr-v119-wang20k}. This point helps to maximize the capacity of each hash bit \cite{shen2018unsupervised}, preserving as much information of the data as possible.
% Too many hash collisions imply a small hash range ($|\{b_i\}_{i=1}^N|$), which not only neglects fine-grained differences, but also result in unstable retrieval performances. 

Finally, the loss of consistency learning is formulated as 
\begin{equation}
    \mathcal{L} =\mathcal{L}_{PSC} + \mathcal{L}_{CSC} + \eta  \mathcal{L}_{CC}
    \label{equ:loss}
\end{equation}
in which $\eta$ is a balance coefficient to balance different consistency loss. However, the $sign(\cdot)$ is in-differentiable at zero and the derivation of it will be zero for every non-zero input, with the result that the parameters of the hashing model will not be updated by the back-propagation algorithm when minimizing the Equation \ref{equ:loss}. Thus, we use $tanh(\cdot)$ to approximate the sign function and generate the approximate hash code $v_i^{(m)} = tanh(G(x_i^{(m)}))$ to replace $b_i^{(m)}$ in loss function. 
Our loss function is optimized by the mini-batch standard stochastic gradient descent (SGD) method. The whole learning procedure are shown in Algorithm 1.

\begin{algorithm}
\caption{CIMON's Training Algorithm }
\label{alg1}
\begin{algorithmic}[1]
\REQUIRE Training images $\mathcal{X}=\{x_{i}\}_{i=1}^{N}$; Code length $L$;\\
% \quad \ \ Similar criteria threshold: $t$,;\\
% \quad \ \ Confidence thresholds: $d_l$, $d_r$;
\ENSURE Parameters $\Theta$ for the neural network $G(\cdot)$;\\
\qquad\ Hash codes $B=\{b_i\}_{i=1}^N$ for training images.
\STATE Generate two transformed images via data augmentation for each image: $\mathcal{X}^{(1)}$ and $\mathcal{X}^{(2)}$;
\FOR{$m=1,2$}
\STATE Get pre-train features of $\mathcal{X}^{(m)}$ through $F(\cdot)$;
\STATE Construct the pseudo-graph $S^{(m)}$ by Equation \ref{equ:similarity_1};
\STATE Perform global refinement to obtain refined pseudo-graph $\hat{S}^{(m)}$ by Equation \ref{equ:similarity_2};
\STATE Construct the confidence matrix $W^{(m)}$ by Equation \ref{equ:confidence};
\ENDFOR
\REPEAT
\STATE Sample $M$ images from $\mathcal{X}$ and obtain their augmentation to construct a mini-batch;
\STATE Calculate loss function by Equation \ref{equ:loss};
\STATE Update parameters of $G(\cdot)$ through back propagation;
\UNTIL convergence
\STATE Generate image hash codes B
\end{algorithmic}
\end{algorithm}

\section{Experiments}
\begin{table*}[h]
\small
\centering
	\begin{tabular*}{1\textwidth}{@{\extracolsep{\fill}}c|cccc|cccc|cccc@{}}
		\hline
		\multirow{2}{*}{Methods} & \multicolumn{4}{c|}{FLICKR25K} & \multicolumn{4}{c|}{CIFAR-10} & \multicolumn{4}{c}{NUS-WIDE} \\

		& 16bits & 32bits & 64bits &128bits & 16bits & 32bits & 64bits &128bits & 16bits & 32bits & 64bits &128bits\\
		\hline
		ITQ & 0.6492 & 0.6518 & 0.6546 & 0.6577& 0.1942 & 0.2086 & 0.2151 & 0.2188 & 0.5270 & 0.5241 & 0.5334 & 0.5398 \\
% 		SH & 0.6091 & 0.6105 & 0.6033 & 0.6014 & 0.1605 & 0.1583 & 0.1509 & 0.1538 & 0.4350 & 0.4129 & 0.4062 & 0.4100\\
		DSH & 0.6452 & 0.6547 & 0.6551 & 0.6557 & 0.1616 & 0.1876 & 0.1918 & 0.2055 & 0.5123 & 0.5118 & 0.5110 & 0.5267\\
		SpH & 0.6119 & 0.6315 & 0.6381 & 0.6451 & 0.1439 & 0.1665 & 0.1783 & 0.1840 & 0.4458 & 0.4537  & 0.4926 & 0.5000\\
		SGH & 0.6362 & 0.6283 & 0.6253 & 0.6206 & 0.1795 & 0.1827 & 0.1889 & 0.1904 & 0.4994 & 0.4869 & 0.4851 & 0.4945\\
		DeepBit &0.5934 & 0.5933 & 0.6199 & 0.6349 & 0.2204 & 0.2410 & 0.2521 & 0.2530 & 0.3844 & 0.4341 & 0.4461 & 0.4917\\
		SSDH & 0.7240 & 0.7276 & 0.7377 & 0.7343 & 0.2568 & 0.2560 & 0.2587 & 0.2601 & 0.6374 & 0.6768 & 0.6829 & 0.6831\\
		DistillHash & - & -&-&-&0.2844&0.2853&0.2867&0.2895&-&-&-&-\\
		CUDH & 0.7332 & 0.7426 & 0.7549 & 0.7561 & 0.2856&0.2903 &0.3025&0.3000&0.6996&0.7222&0.7451&0.7418\\
		MLS$^3$RDUH &0.7587 &0.7754 &0.7870 &0.7927 &0.2876 &0.2962 & 0.3139 &0.3117 &0.7056 &0.7384 &0.7629 &0.7818 \\
		\hline
		CIMON  &\textbf{0.8049} & \textbf{0.8195} & \textbf{0.8281} & \textbf{0.8321} & \textbf{0.4506} & \textbf{0.4723} & \textbf{0.4944} & \textbf{0.4981} & \textbf{0.7883} & \textbf{0.8060} & \textbf{0.8214} & \textbf{0.8243} \\
		\hline
	\end{tabular*}
	\caption{MAP for different methods on FLICKR25K, CIFAR-10 and NUS-WIDE.\label{tab:1}}
\end{table*}
We conduct extensive experiments on three popular benchmark datasets to evaluate our CIMON by comparisons with various unsupervised hashing methods.
\subsection{Datasets and Setup}
\textbf{FLICKR25K} \cite{huiskes2008mir} contains 25,000 images labeled by some of the 24 categories. We randomly sample 2,000 images as the query set and use the other images as the retrieval set. 10000 images are randomly selected from the retrieval set as the training set. 
\textbf{CIFAR-10} \cite{krizhevsky2009learning} contains 60000 images of ten different categories. We randomly sample 1,000 images as the query set for each class, and take the rest as the retrieval set. We sample 500 images per class in the retrieval set as the training set. 
\textbf{NUSWIDE} \cite{chua2009nus} contains 269,648 images of 81 concepts. Here, we use the subset that contains the 10 most popular concepts. We randomly select 5,000 images as the query set and the remaining images make up the retrieval set. 5000 images randomly sampled from the retrieval set serve as the training set. 

Our CIMON is compared with various state-of-the-art unsupervised hashing methods including both traditional methods and deep learning methods. Traditional methods include ITQ \cite{gong2012iterative}, 
% SH \cite{weiss2009spectral}, 
DSH \cite{jin2013density}, SpH \cite{heo2012spherical} and SGH \cite{dai2017stochastic}. Deep unsupervised hashing methods include DeepBits \cite{lin2016learning}, SSDH \cite{yang2018semantic}, DistillHash \cite{yang2019distillhash}, CUDH \cite{2019Clustering}, and MLS$^3$RUDH
% \footnote{Codes provided by the authors}
\cite{ijcai2020-479}. For deep learning-based methods, we use raw pixels as inputs. For traditional methods, we extract $4096$-dimensional feature vectors by the VGG-F model which is pre-trained on ImageNet for fair comparison.
% For fair comparisons, all the methods are reported with identical training and test sets. Additionally, the shallow methods are evaluated with the same deep features as the ones we are using.

As for evaluation, the ground-truth similarity information is obtained according to the ground-truth image labels. Specifically, two data points are considered similar if they share the same label or share at least one common label. The retrieval quality is evaluated by Mean Average Precision (MAP), Precision-Recall curve and Top N precision curve. MAP is widely-used to evaluate the retrieval accuracy. Given a query and a list of $R$ ranked retrieval results, the average precision (AP) for the given query can be obtained. MAP is defined to be the average of APs for all queries. For experiments on datasets FLICKR25K and NUSWIDE, we set $R$ as 5000. For CIFAR-10, we set $R$ as $50000$. Precision-recall curve can reveal the precision at different recall levels, which is a good indicator of overall performance. As a popular way of the performance visualization, Top N precision curve is the precision curve in regard to the top N retrieved instances.

In our implementation, we optimize our model by mini-batch SGD with momentum. 
The mini-batch size is set to 24. The learning rate is fixed at $0.001$. For all three datasets, training images are resized to 224 $\times$ 224 as inputs. 
Data augmentation we adopt includes random cropping and resizing, rotation, cutout, color distortion and Gaussian blur. As two introduced hyper-parameters, $\eta$ and the number of clusters $K$ in spectral clustering are set to $0.3$ and $70$ as default. 

% \subsection{Evaluation}
% The ground-truth similarity information for evaluation is constructed from the ground-truth image labels: two data points are considered similar if they share the same label (for CIFAR-10, STL-10 and Tiny-ImageNET) or share at least one common label (for FLICKR25K and NUSWIDE).

% The retrieval quality are evaluated by the following three evaluation metrics: Mean Average Precision (MAP), Precision-Recall curve and Top N precision curve. MAP is a widely-used criteria to evaluate retrieval accuracy. Given a query and a list of $R$ ranked retrieval results, the average precision (AP) for the given query can be computed. MAP is defined as the average of APs for all queries. For datasets FLICKR25K and NUSWIDE, we set $R$ as $5000$ for the experiments. For CIFAR-10, STL-10 and Tiny-Imagenet, $R$ is set to 
% the number of images in the database, $50000$, $5000$ and $10000$ respectively. Precision-recall curve reveals the precision at different recall levels and is a good indicator of overall performance. In addition, Top N precision curve, which is the precision curve with respect to the top $K$ retrieved instances, also visualizes the performance from a different perspective. 
\subsection{Experimental Results}

\begin{figure}[!h]
\centering
\includegraphics[width=3.4in,keepaspectratio=true]{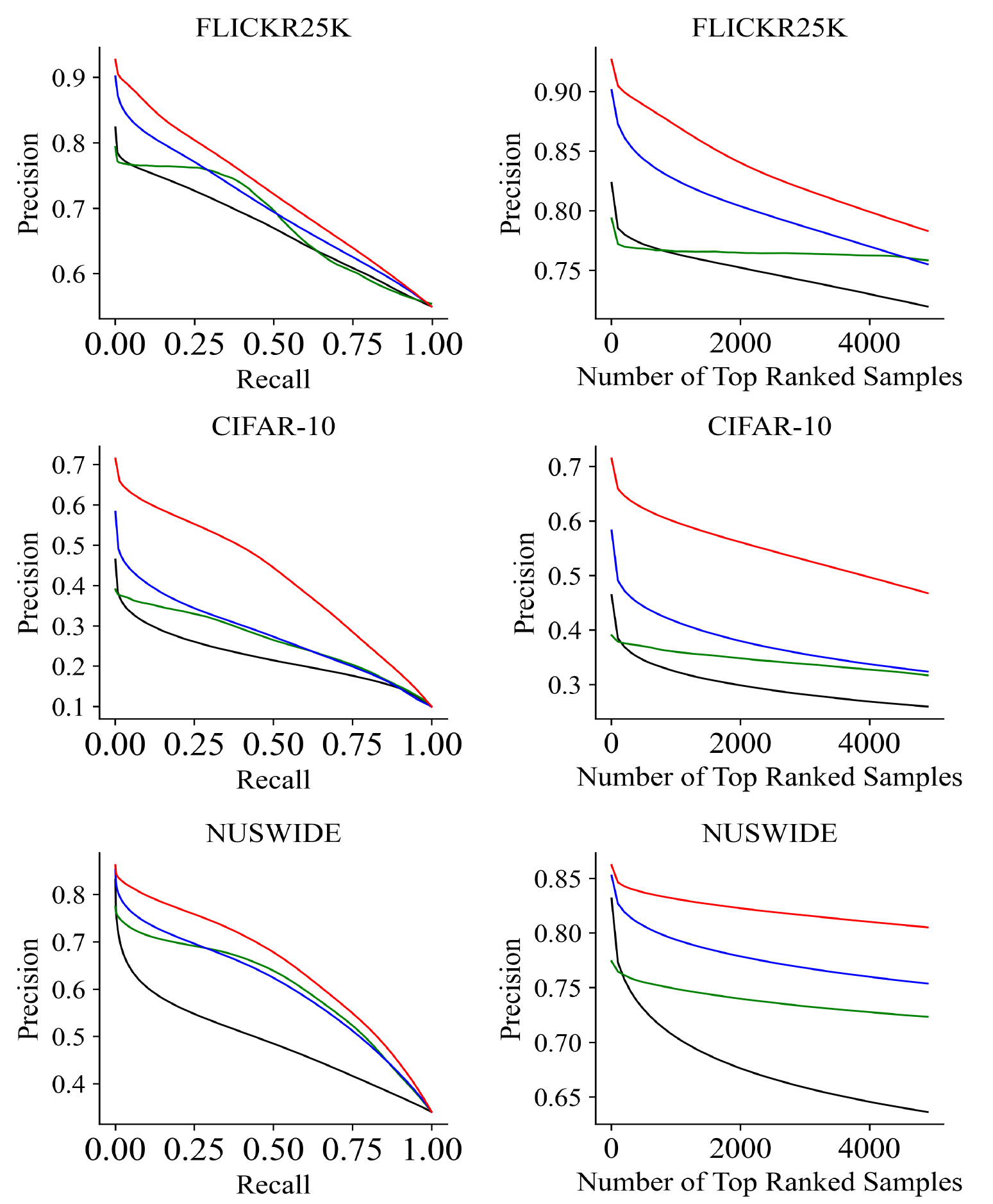}
\caption{The first column plots the precision-recall curves, and the second column plots the Top-$N$ precision curves. ($L=128$, {\color{black}---} SSDH, {\color{green}---} CUDH, {\color{blue}---} MLS$^3$RDUH, {\color{red}---} CIMON) \label{fig:prc}}
\end{figure}

 The MAPs of different methods on datasets FLICKER25K, CIFAR-10 and NUSWIDE with hash code lengths varying from 16 to 128 are shown in Table \ref{tab:1}. The following observations can be derived: 1) Deep learning-based algorithms overall perform better than the traditional methods, which shows that the strong representation-learning ability of deep learning helps to improve the performance of unsupervised hashing methods. 2) The methods that reconstruct semantic similarity structure with global information (CUDH, MLS$^3$RUDH) perform better than other deep unsupervised hashing methods, which indicates that semantic similarity reconstructed only by local information (i.e. pairwise distance of features) is inaccurate and unreliable. 3) We can find that CIMON has a significant improvement over the previous the-state-of-art MLS$^3$RUDH in all cases by a large margin. Specifically, the improvements of our model over the best baseline are 5.51\%, 58.4\% and 8.39\% for average MAP on datasets FLICKER25K, CIFAR-10 and NUS-WIDE respectively, which shows the superiority of our model.
We plot the precision-recall curves of SSDH, CUDH, MLS$^3$RDUH and CIMON on three datasets in the first column of Figure \ref{fig:prc}. We find that the curve of CIMON is always on top of the other three models' curves, which demonstrates that the hash codes obtained by CIMON are also more appropriate for hash table lookup search strategy. The second column of Figure \ref{fig:prc} shows that the Top-$N$ precision curves of these four models on the same datasets. Our CIMON significantly outperforms the comparison methods by large margins. Since the precision curves are based on the ranks of Hamming distance, CIMON achieves superior performance under Hamming ranking-based evaluations. 
% \begin{table}[h]
% % \scriptsize
% \small
% \centering
% \caption{MAP for different deep hashing methods on STL-10.\label{tab:2}}
% 	\begin{tabular*}{0.47\textwidth}{@{\extracolsep{\fill}}c|cccc@{}}
% 		\hline
% 		\multirow{2}{*}{Methods} & \multicolumn{4}{c}{STL-10} \\
% 		 & 16bits & 32bits & 64bits &128bits \\
% 		\hline
% 		SSDH & 0.4447 & 0.4558 & 0.4612 & 0.4497 \\
% 		CUDH &0.5010 & 0.5383 & 0.6378  & 0.6108 \\
% 		MLS$^3$RDUH & 0.6000 & 0.6266 & 0.6595  & 0.6646 \\		
% 		CIMON &\textbf{0.7243} & \textbf{0.7595} & \textbf{0.7717} & \textbf{0.7721} \\
% 		\hline
% 	\end{tabular*}
% \end{table}
% To fully explore the performance of our model on single-labeled dataset, we further study the performance of our model on STL-10 compared with three best baselines, and get the consistent results, which are shown in Table \ref{tab:2}.
% Table \ref{tab:2} shows the MAPs for different methods on datasets STL-10 and Tiny-ImageNET.
% Our model significantly surpasses other methods by a large margin on these benchmarks under all code length. Concretely, 
% For STL-10 dataset, our model outperforms the best baseline about 18.70\%. As for Tiny-ImageNet dataset, it contains $200$ categories, which implies a hard task for image retrieval. Our model still outperforms all the baselines on this dataset, which further demonstrates the superiority of our model.

% Besides, Top-$N$ precision curves for the four best methods with the hash code length of 128 are shown for a more comprehensive comparison.

\begin{figure}[!h]
\centering
\includegraphics[width=3.4in,keepaspectratio=true]{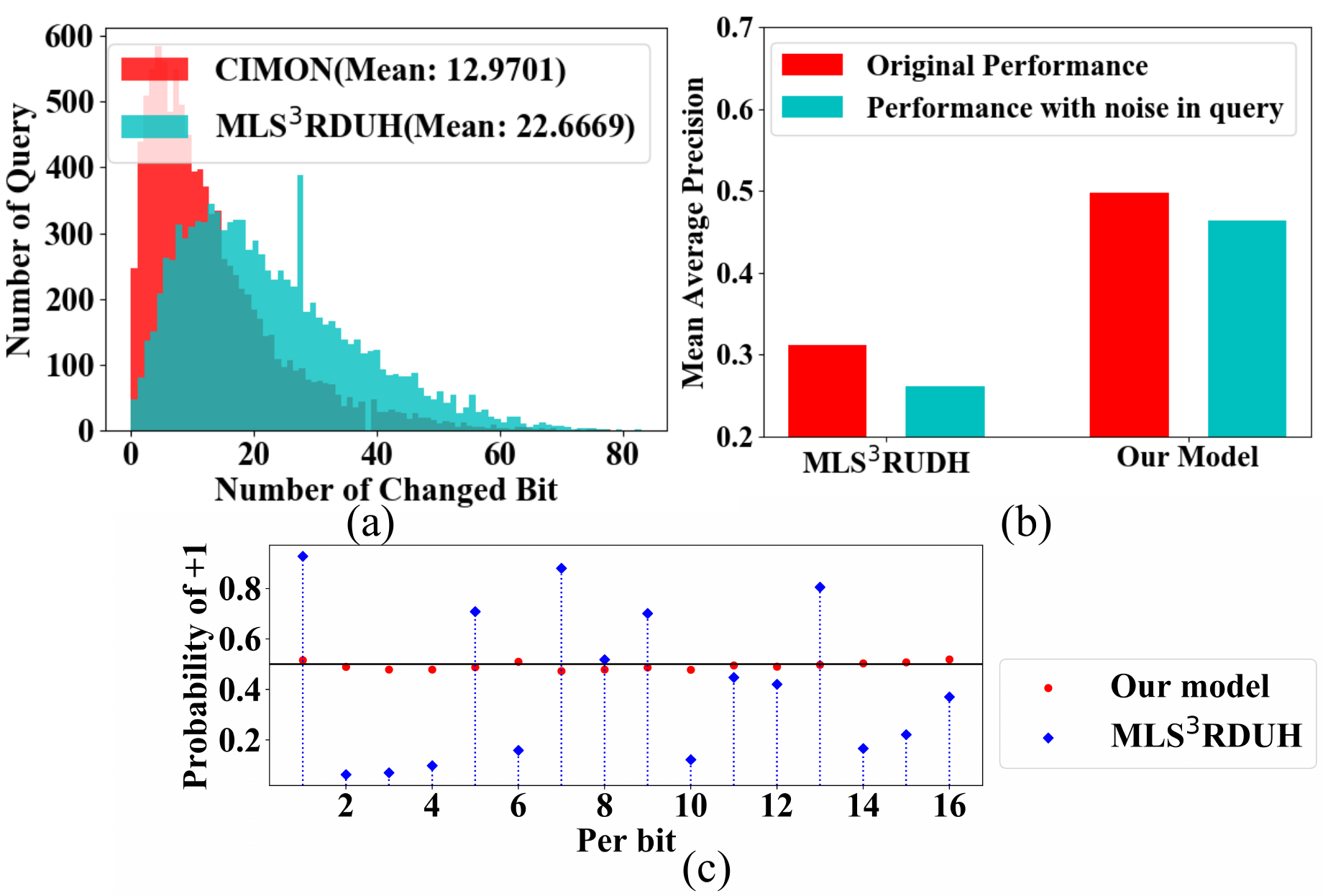}
\caption{(a) Distribution of changed bits number after adding noise in query images on CIFAR-10 for MLS$^3$RUDH and CIMON, respectively. (b) MAP results before and after adding noise to MLS$^3$RUDH and CIMON. (c) The per-bit probability of code +1 over all images for 16 bits on CIFAR-10. }\label{ro}
\end{figure}

 To demonstrate the robustness of CIMON, we add perturbation or transformation noise to the query set, which doesn't break the semantic information. Figure \ref{ro}(a) shows the distribution of changed bits number before and after adding noise in query images for MLS$^{3}$RUDH and our model. 
%It is observed that our model has better transformation invariant compared with the baseline. The MAP after the noise attack also decreases less compared with the baseline in Figure 4 (b).
It is observed that the mean of changed bits number of CIMON is significantly smaller than that of MLS$^{3}$RUDH, which implies that CIMON can learn more disturb-invariant hash codes. The MAP of CIMON also decreases less after the noise attack compared with the baseline in Figure \ref{ro}(b). Moreover, CIMON is able to generate informative hash bits because hash bits distribution of CIMON approximates a uniform distribution, making good use of full bit capacity in Figure \ref{ro}(c). 

% \noindent{\bf{Fewer Collisions}}. 
% Figure \ref{ro} (c) shows the number of collisions in the query dataset for our model with contrastive consistency or not. It can be found that our full model suffer the fewer collisions, which  implies that contrastive consistency helps to generate high-quality hash codes with fewer collisions.  

\begin{table}[!h] 
\footnotesize
\centering
	\begin{tabular}{@{\extracolsep{-0.1cm}}c|cccc|cccc@{}}
		\hline
		\multirow{2}{*}{} & \multicolumn{4}{c|}{Correlations} & \multicolumn{4}{c}{Results} \\
		& GR & CW & SC & CC  & 16bits & 32bits & 64bits &128bits\\
		\hline
		$M_1$  &  &  &  &  &  0.1907 & 0.2253 & 0.2428 & 0.2592\\
		$M_2$  & $\surd$ &  &  &  &0.2101 & 0.2552 & 0.2640 & 0.2796\\
		$M_3$ & $\surd$ & $\surd$ &  &  & 0.3251 & 0.3260 & 0.3354 & 0.3452\\
		$M_4$ & $\surd$ & $\surd$ & $\surd$ &  & 0.3815 &0.4315  &0.4399  &0.4437 \\
		$M_5$ & $\surd$ & $\surd$ & $\surd$ & $\surd$ & \textbf{0.4506} & \textbf{0.4723} & \textbf{0.4944} & \textbf{0.4981} \\
		\hline
	\end{tabular}
\caption{Ablation analysis on CIFAR-10. GR, CW, SC and CC correspond to Global Refinement, Confidence-based Weight, Semantic Consistency and Contrastive Consistency, respectively.\label{tab:ab}}
\end{table}

\noindent{\bf{Ablation Study}} We investigate the effectiveness of various correlations in Table \ref{tab:ab}. $M_1$ uses the local similarity structure as guiding information, and trains the hashing network with the degraded loss following \cite{yang2018semantic}. The difference between $M_2$ and $M_1$ lies in whether to use the global refinement or not. It can be seen that $M_2$ surpasses $M_1$ significantly, demonstrating the effectiveness of global refinement for reconstructing the accurate similarity graph. After considering the confidence of semantic similarity, $M_3$ achieves better results than $M_2$ because the refined similarity-graph is still noisy and $M_3$ further accounts for the variations in confident and unconfident pairs, which eases the effect of false similarity signals and enlarges the effect of highly confident signals in the similarity graph. $M_4$ makes use of the data augmentation and our novel semantic consistency loss function. We can see that $M_4$ performs much better than $M_3$, which demonstrates the strength of data augmentation and our well-designed semantic consistency loss. By comparing the results of $M_5$ and $M_4$, we can see that the contrastive consistency can further improve the performance of our model. 

\begin{figure}[!h]
\centering
\includegraphics[width=3.4in,keepaspectratio=true]{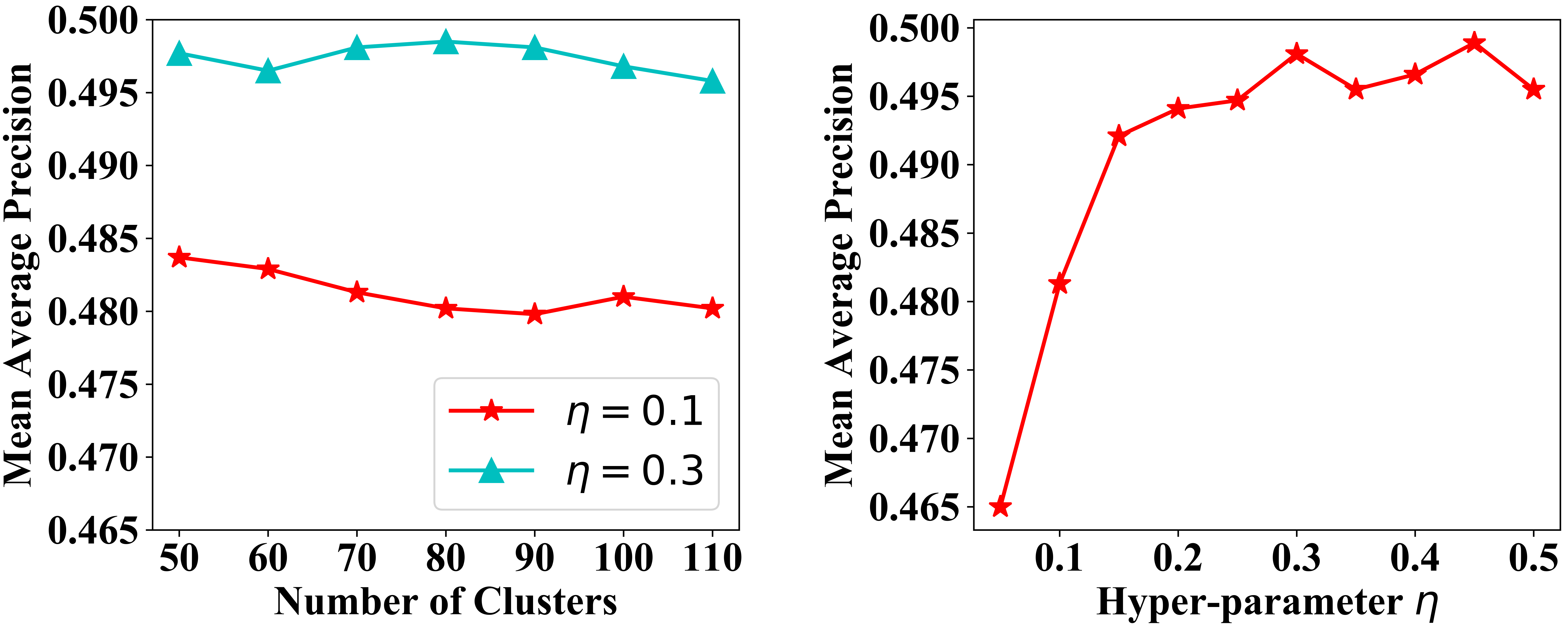}
\caption{Sensitivity of $K$ and $\eta$ with 128-bits on CIFAR-10\label{fig:sen2}}
\end{figure}

\noindent{\bf{Parameter Sensitivity}} We study the influence of balance coefficient $\eta$ and the number of clusters $K$ in Figure \ref{fig:sen2}. We first fix $\eta$ to $0.1$ and $0.3$ and evaluate the MAP by varying $K$ from $50$ to $110$. The performance of our model is not sensitive to the number of clusters in the range of $[50,110]$ and we can set $K$ as any values in that interval. Furthermore, we show the MAP by varying $\eta$ from $0.05$ to $0.5$ with $K$ fixed to $70$. The MAP of our model first increases and then keeps at a relatively high level. The result is not sensitive to $\eta$ in the range of $[0.2,0.5]$. Then for the proposed model, K and $\eta$ are set to $70$ and $0.3$ respectively.

\begin{figure}[ht]
\centering
\includegraphics[width=3.3in,keepaspectratio=true]{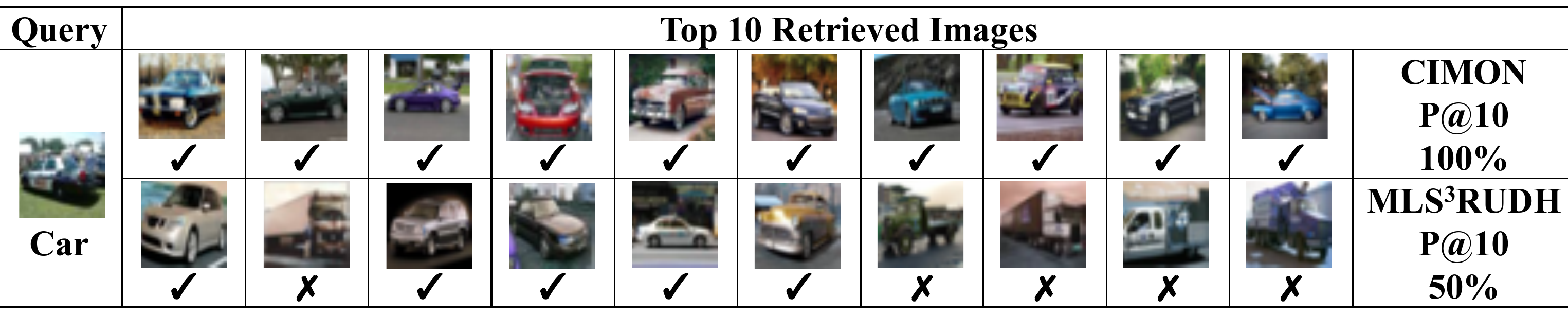}
\caption{Examples of the top 10 images and Precision@10}
\label{fig:visual}
\end{figure}
% \noindent{\bf{Top 10 retrieval results}} 
\noindent{\bf{Visualization}} In Figure \ref{fig:visual}, we visualize the top 10 returned images of our model and the best baseline for the query image of CIFAR-10, which demonstrates that our model can retrieve much more relevant and user-desired images.
% \subsection{Visualization}
% \noindent{\bf{Visualization of Hash Codes by t-SNE.}} The t-SNE visualization \cite{cao2018deep} of the hash codes learned by our model and the best baseline MLS$^3$RUDH on CIFAR-10 are shown in Figure \ref{fig:tsne}. It is observed that the hash codes generated by our model show more discriminative structures where the hash codes in different categories are better separated, compared with the result of MLS$^3$RUDH. This result verifies that the hash codes generated by our model preserve semantic similarity much better and enable more effective image retrieval. 
% \begin{figure}[h]
% 	\centering
% 	\subfigure[MLS$^3$RDUH]{
% 	\begin{minipage}[t]{0.45\linewidth}
% 	\centering
% 	\includegraphics[width=4cm]{image//S3.jpg}
% 	\end{minipage}
% 	}
% 	\subfigure[Our model]{
% 	\begin{minipage}[t]{0.45\linewidth}
% 	\centering
% 		\includegraphics[width=4cm]{image//Our_model.jpg}
% 	\end{minipage}
% 	}
% \caption{The t-SNE visualization of hash codes on CIFAR-10.}
% \label{fig:ts3.4in
% \end{figure}
% \begin{figure}[ht]
% \centering
% \includegraphics[width=3.4in,keepaspectratio=true]{image//view.pdf}
% \caption{Examples of the top 10 images and Precision@10 on CIFAR-10}
% \label{fig:visual}
% \end{figure}
% % \noindent{\bf{Top 10 retrieval results}} 
% In Figure \ref{fig:visual}, we visualize the top 10 returned images of our model and the best baseline for three query images of CIFAR-10, which demonstrates that our model can retrieve much more relevant and user-desired images.

\section{Conclusion}
Here we propose a novel deep hashing method named CIMON, which generates reliable semantic information by comprehensive similarity mining from local and global views. Then a novel consistency loss function from the view of semantic matching and contrastive learning is proposed to optimize the hashing model by incorporating the semantic information into the training process. 
Extensive experiments reveal that CIMON boosts the state-of-the-art unsupervised hashing schemes in both image retrieval and robustness.

\section*{Acknowledgements}
This work was supported by The National Key Research and Development Program of China (No.2016YFA0502303) and the National Natural Science Foundation of China (No.31871342).

\newpage
\bibliographystyle{named}
\bibliography{ms}

\begin{thebibliography}{}

\bibitem[\protect\citeauthoryear{Boussaha \bgroup \em et al.\egroup
  }{2019}]{boussaha2019multi}
Basma El~Amel Boussaha, Nicolas Hernandez, Christine Jacquin, and Emmanuel
  Morin.
\newblock Multi-level context response matching in retrieval-based dialog
  systems.
\newblock In {\em Proceedings of the 7th edition of the Dialog System
  Technology Challenges Workshop at AAAI}, 2019.

\bibitem[\protect\citeauthoryear{Chen \bgroup \em et al.\egroup
  }{2020}]{chen2020simple}
Ting Chen, Simon Kornblith, Mohammad Norouzi, and Geoffrey Hinton.
\newblock A simple framework for contrastive learning of visual
  representations.
\newblock In {\em ICML}, 2020.

\bibitem[\protect\citeauthoryear{Chua \bgroup \em et al.\egroup
  }{2009}]{chua2009nus}
Tat-Seng Chua, Jinhui Tang, Richang Hong, Haojie Li, Zhiping Luo, and Yantao
  Zheng.
\newblock Nus-wide: a real-world web image database from national university of
  singapore.
\newblock In {\em Proceedings of the ACM international Conference on Image and
  Video Retrieval}, page~48, 2009.

\bibitem[\protect\citeauthoryear{Dai \bgroup \em et al.\egroup
  }{2017}]{dai2017stochastic}
Bo~Dai, Ruiqi Guo, Sanjiv Kumar, Niao He, and Le~Song.
\newblock Stochastic generative hashing.
\newblock In {\em ICML}, pages 913--922, 2017.

\bibitem[\protect\citeauthoryear{Dosovitskiy \bgroup \em et al.\egroup
  }{2014}]{dosovitskiy2014discriminative}
Alexey Dosovitskiy, Jost~Tobias Springenberg, Martin Riedmiller, and Thomas
  Brox.
\newblock Discriminative unsupervised feature learning with convolutional
  neural networks.
\newblock In {\em NIPS}, pages 766--774. 2014.

\bibitem[\protect\citeauthoryear{Fan \bgroup \em et al.\egroup
  }{2020}]{fan20deep}
Lixin Fan, Kam~Woh Ng, Ce~Ju, Tianyu Zhang, and Chee~Seng Chan.
\newblock Deep polarized network for supervised learning of accurate binary
  hashing codes.
\newblock In {\em IJCAI}, pages 825--831, 2020.

\bibitem[\protect\citeauthoryear{Gong \bgroup \em et al.\egroup
  }{2012}]{gong2012iterative}
Yunchao Gong, Svetlana Lazebnik, Albert Gordo, and Florent Perronnin.
\newblock Iterative quantization: A procrustean approach to learning binary
  codes for large-scale image retrieval.
\newblock {\em IEEE Transactions on Pattern Analysis and Machine Intelligence},
  35(12):2916--2929, 2012.

\bibitem[\protect\citeauthoryear{Gu \bgroup \em et al.\egroup
  }{2019}]{2019Clustering}
Yifan Gu, Shidong Wang, Haofeng Zhang, Yazhou Yao, and Li~Liu.
\newblock Clustering-driven unsupervised deep hashing for image retrieval.
\newblock {\em Neurocomputing}, 368:114--123, 2019.

\bibitem[\protect\citeauthoryear{Hadsell \bgroup \em et al.\egroup
  }{2006}]{hadsell2006dimensionality}
Raia Hadsell, Sumit Chopra, and Yann LeCun.
\newblock Dimensionality reduction by learning an invariant mapping.
\newblock In {\em CVPR}, pages 1735--1742, 2006.

\bibitem[\protect\citeauthoryear{He \bgroup \em et al.\egroup
  }{2020}]{he2020momentum}
Kaiming He, Haoqi Fan, Yuxin Wu, Saining Xie, and Ross Girshick.
\newblock Momentum contrast for unsupervised visual representation learning.
\newblock In {\em CVPR}, pages 9729--9738, 2020.

\bibitem[\protect\citeauthoryear{Heo \bgroup \em et al.\egroup
  }{2012}]{heo2012spherical}
Jae-Pil Heo, Youngwoon Lee, Junfeng He, Shih-Fu Chang, and Sung-Eui Yoon.
\newblock Spherical hashing.
\newblock In {\em CVPR}, pages 2957--2964, 2012.

\bibitem[\protect\citeauthoryear{Huiskes and Lew}{2008}]{huiskes2008mir}
Mark~J Huiskes and Michael~S Lew.
\newblock The mir flickr retrieval evaluation.
\newblock In {\em ACMMM}, pages 39--43, 2008.

\bibitem[\protect\citeauthoryear{Jin \bgroup \em et al.\egroup
  }{2013}]{jin2013density}
Zhongming Jin, Cheng Li, Yue Lin, and Deng Cai.
\newblock Density sensitive hashing.
\newblock {\em IEEE transactions on cybernetics}, 44(8):1362--1371, 2013.

\bibitem[\protect\citeauthoryear{Krizhevsky \bgroup \em et al.\egroup
  }{2009}]{krizhevsky2009learning}
Alex Krizhevsky, Geoffrey Hinton, et~al.
\newblock Learning multiple layers of features from tiny images.
\newblock 2009.

\bibitem[\protect\citeauthoryear{Lin \bgroup \em et al.\egroup
  }{2016}]{lin2016learning}
Kevin Lin, Jiwen Lu, Chu-Song Chen, and Jie Zhou.
\newblock Learning compact binary descriptors with unsupervised deep neural
  networks.
\newblock In {\em CVPR}, pages 1183--1192, 2016.

\bibitem[\protect\citeauthoryear{Luo \bgroup \em et al.\egroup
  }{2020}]{luo2020survey}
Xiao Luo, Chong Chen, Huasong Zhong, Hao Zhang, Minghua Deng, Jianqiang Huang,
  and Xiansheng Hua.
\newblock A survey on deep hashing methods.
\newblock {\em arXiv preprint arXiv:2003.03369}, 2020.

\bibitem[\protect\citeauthoryear{Luo \bgroup \em et al.\egroup
  }{2021}]{luo2021}
Xiao Luo, Zeyu Ma, Daqing Wu, Huasong Zhong, Jinwen Ma, and Minghua Deng.
\newblock Deep unsupervised hashing by distilled smooth guidance.
\newblock In {\em ICME}, 2021.

\bibitem[\protect\citeauthoryear{Shen \bgroup \em et al.\egroup
  }{2018}]{shen2018unsupervised}
Fumin Shen, Yan Xu, Li~Liu, Yang Yang, Zi~Huang, and Heng~Tao Shen.
\newblock Unsupervised deep hashing with similarity-adaptive and discrete
  optimization.
\newblock {\em IEEE transactions on pattern analysis and machine intelligence},
  40(12):3034--3044, 2018.

\bibitem[\protect\citeauthoryear{Simonyan and
  Zisserman}{2015}]{simonyan2014very}
K.~Simonyan and A.~Zisserman.
\newblock Very deep convolutional networks for large-scale image recognition.
\newblock In {\em ICLR}, May 2015.

\bibitem[\protect\citeauthoryear{Tu \bgroup \em et al.\egroup
  }{2020}]{ijcai2020-479}
Rong-Cheng Tu, Xian-Ling Mao, and Wei Wei.
\newblock Mls3rduh: Deep unsupervised hashing via manifold based local semantic
  similarity structure reconstructing.
\newblock In {\em IJCAI}, pages 3466--3472, 2020.

\bibitem[\protect\citeauthoryear{Wang and Isola}{2020}]{pmlr-v119-wang20k}
Tongzhou Wang and Phillip Isola.
\newblock Understanding contrastive representation learning through alignment
  and uniformity on the hypersphere.
\newblock In {\em ICML}, pages 9929--9939, 2020.

\bibitem[\protect\citeauthoryear{Wang \bgroup \em et al.\egroup
  }{2020}]{wang2020set}
Weiwei Wang, Yuming Shen, Haofeng Zhang, Yazhou Yao, and Li~Liu.
\newblock Set and rebase: determining the semantic graph connectivity for
  unsupervised cross modal hashing.
\newblock In {\em IJCAI}, pages 853--859, 2020.

\bibitem[\protect\citeauthoryear{Wu \bgroup \em et al.\egroup
  }{2018}]{wu2018unsupervised}
Zhirong Wu, Yuanjun Xiong, Stella~X Yu, and Dahua Lin.
\newblock Unsupervised feature learning via non-parametric instance
  discrimination.
\newblock In {\em CVPR}, pages 3733--3742, 2018.

\bibitem[\protect\citeauthoryear{Wu \bgroup \em et al.\egroup
  }{2019}]{wu2019deep}
Jianlong Wu, Keyu Long, Fei Wang, Chen Qian, Cheng Li, Zhouchen Lin, and
  Hongbin Zha.
\newblock Deep comprehensive correlation mining for image clustering.
\newblock In {\em CVPR}, pages 8150--8159, 2019.

\bibitem[\protect\citeauthoryear{Xie \bgroup \em et al.\egroup
  }{2020}]{xie2020label}
Yanzhao Xie, Yu~Liu, Yangtao Wang, Lianli Gao, Peng Wang, and Ke~Zhou.
\newblock Label-attended hashing for multi-label image retrieval.
\newblock In {\em IJCAI}, pages 955--962, 2020.

\bibitem[\protect\citeauthoryear{Yang \bgroup \em et al.\egroup
  }{2018}]{yang2018semantic}
Erkun Yang, Cheng Deng, Tongliang Liu, Wei Liu, and Dacheng Tao.
\newblock Semantic structure-based unsupervised deep hashing.
\newblock In {\em IJCAI}, pages 1064--1070, 2018.

\bibitem[\protect\citeauthoryear{Yang \bgroup \em et al.\egroup
  }{2019}]{yang2019distillhash}
Erkun Yang, Tongliang Liu, Cheng Deng, Wei Liu, and Dacheng Tao.
\newblock Distillhash: Unsupervised deep hashing by distilling data pairs.
\newblock In {\em CVPR}, pages 2946--2955, 2019.

\end{thebibliography}

\end{document}